\pdfoutput=1

\documentclass[11pt]{article}

\usepackage[final]{acl}

\usepackage{times}
\usepackage{latexsym}

\usepackage[T1]{fontenc}

\usepackage[utf8]{inputenc}

\usepackage{microtype}

\usepackage{inconsolata}

\usepackage{graphicx}
\usepackage{booktabs}
\usepackage{tablefootnote}
\usepackage{multicol, multirow,makecell}

\usepackage{amsmath}
\usepackage{xcolor}
\usepackage{colortbl}

\definecolor{OliveGreen}{rgb}{0,0.6,0}
\newcolumntype{C}[1]{>{\raggedright\arraybackslash}p{#1}}

\title{iTBLS: A Dataset of Interactive Conversations Over Tabular Information}

\author{
    Anirudh Sundar\textsuperscript{1} \and
    Christopher Richardson\textsuperscript{2} \thanks{Work done while at Georgia Tech} \and
    Adar Avsian\textsuperscript{1} \and 
    \textbf{Larry Heck}\textsuperscript{1} \\
    \textsuperscript{1} Georgia Institute of Technology, USA \\
    \textsuperscript{2} Google Inc., USA \\
    \texttt{asundar34, larryheck@gatech.edu} \\
}

\begin{document}
\maketitle
\begin{abstract}
This paper introduces Interactive Tables (iTBLS), a dataset of interactive conversations that focuses on natural-language manipulation of tabular information sourced from academic pre-prints on ArXiv. The iTBLS dataset consists of three types of tabular tasks -- interpretation, modification, and generation. Interpretation focuses on tabular understanding, modification focuses on manipulating tabular information, and generation focuses on the addition of new natural-language evidence. In addition, the paper presents a novel framework that reformulates tabular operations as question-answering, where an appropriate question is formulated based on the nature of interaction and the question is answered using the user request as evidence. The developed approach results in an improvement on all tasks on a sequence-to-sequence modeling baseline on iTBLS. In addition, the question-answering-based reformulation is applied to datasets from prior work for the text-to-table task where textual paragraphs are summarized into tables. The novel approach results in up to 13\% improvement in Exact-Match accuracy and up to 16\% improvement in BERTScores compared to the prior state-of-the-art. 
\end{abstract}

\section{Introduction}
Recent research on Conversational AI has focused on adding enhanced multi-task capabilities to large language models (LLMs). This research includes building systems capable of situated interactions over structured knowledge sources such as tabular information
\cite{sundar_multimodal_2022}. Automated methods for tabular interpretation, manipulation, and generation empower users by saving time and reducing errors in managing tabular content \cite{kardas-etal-2020-axcell}. Previous studies have focused on individual aspects of tabular data management: representation learning for interpretation tasks like grounded question answering, manipulation for data wrangling, and generation for summarizing textual information independently \cite{nakamura_hybridialogue_2022, sundar-heck-2023-ctbls, fang2024large}. 

The development of situated conversational interactions over tables necessitates a suite of approaches to unify tabular interpretation, modification, and generation in a conversational context. Additionally, an important yet largely unaddressed challenge in interacting with tabular sources is the ability to modify existing tabular content using conversational natural language commands.

To address these challenges, this paper introduces Interactive Tables (\mbox{iTBLS}) \footnote{\url{https://huggingface.co/datasets/avalab/iTBLS}}, a dataset of interactive conversations in English situated in tabular information. iTBLS decomposes the challenge into three distinct tasks: \textit{interpretation}, which involves understanding tabular content within a conversational framework; \textit{modification}, which entails manipulating tabular content through natural language commands; and \textit{generation}, which focuses on integrating new natural language information into existing tables. The tabular information in \mbox{iTBLS} is sourced from scientific articles hosted on arXiv~\footnote{\url{https://arxiv.org}}, an open-access repository of academic preprints. 

Beyond factoid question-answering, \mbox{iTBLS} encompasses tasks such as comparison, determining absolute and relative positions, and mathematical reasoning. Previous research primarily examined procedural command generation for spreadsheets or the alignment of tabular data through LLMs. \mbox{iTBLS} integrates these functionalities into a unified task, enabling the manipulation of existing tables through natural-language commands. 
On tabular generation, while prior work addressed the summarization of natural language paragraphs in a tabular format, \mbox{iTBLS} focuses on generating row or column data conversationally. 

In addition to building iTBLS, this paper develops a novel approach to address tabular operations by reformulating the task as conditional question answering. Furthermore, the question-answering-based reformulation is applied to other datasets introduced in prior work \cite{wu_text--table_2022} and results in better performance in terms of both table-cell accuracy and BERTScore. 

\begin{figure*}[t]
    \centering
    \includegraphics[width=\textwidth]{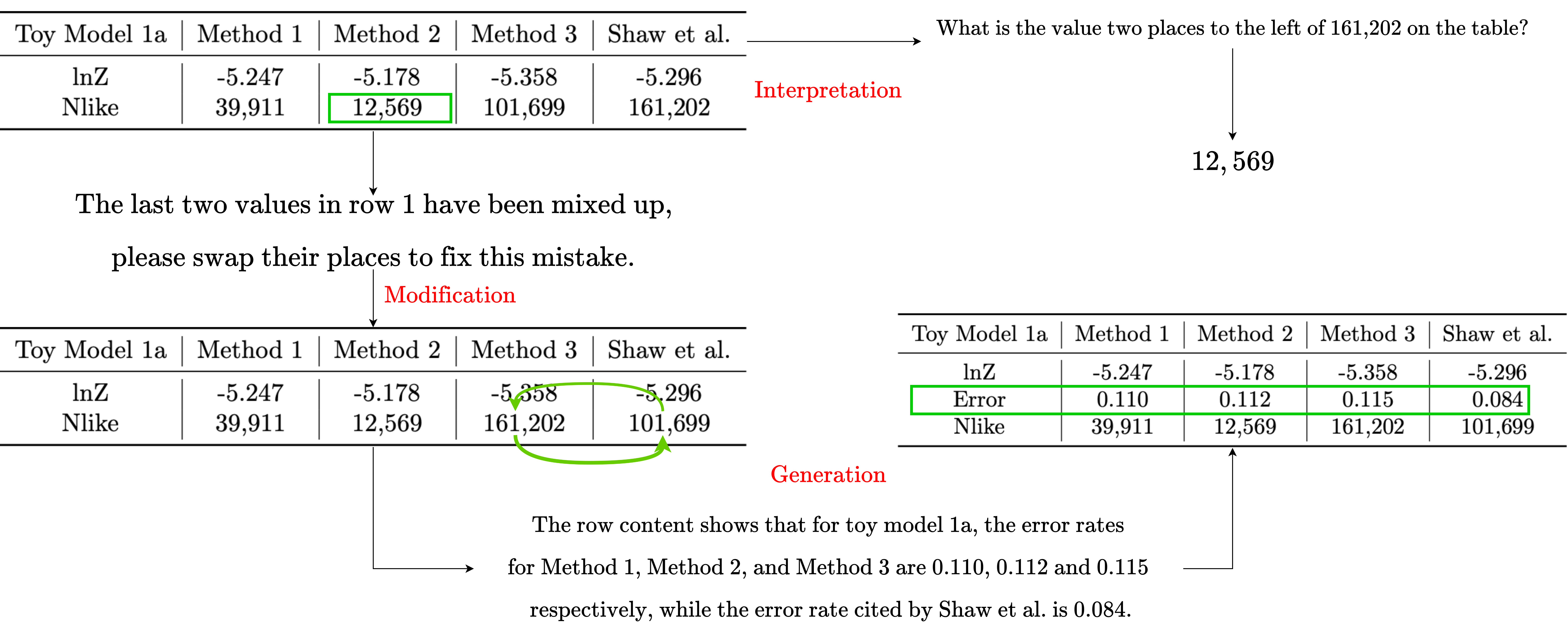}
    \caption{Examples of interactions from the Interactive Tables (\mbox{iTBLS}) dataset.}
    \label{fig:dataset_examples}
\end{figure*}

The contributions of this work are as follows:
\begin{itemize}
    \item Creating \mbox{iTBLS}, a dataset of tabular interactions unifying interpretation, modification, and generation.
    \item Extending prior tabular datasets by collecting information from arXiv
    \item Broadening the scope of interactions to include mathematical reasoning,  natural language manipulation, and natural language expansion.
    \item Introducing a novel approach for table generation tasks through a two-stage reformulation that first identifies the cells to be manipulated and generates a question based on the requested operation, then answers those questions using the user request and the input table as evidence.  
    \item Demonstrating up to 13\% improvement in table-cell accuracy and up to 16\% improvement in BERTScore using the novel approach on the text-to-table task introduced by prior work.  
\end{itemize}

\section{Related Work}
\label{sec:related_work}

A detailed survey of LLMs for tabular data is available in \cite{fang2024large}.
Related work on paired natural-language and tabular data can be broadly classified by the nature of the interaction: tabular interpretation, tabular modification, and tabular generation. 

\subsection{Tabular Interpretation}
Tabular interpretation involves a dialogue turn focused on extracting information from a specific cell in a table, such as identifying a cell satisfying certain criteria. 
Prior research on tabular interpretation focused on grounded question-answering. An important challenge in the collection of such datasets is the availability of large-scale tabular data. Consequently, many tabular datasets are constructed from online resources such as Wikipedia including \textsc{\mbox{WikiTableQuestions}} \cite{pasupat-liang-2015-compositional}, ManyModalQA \cite{hannan2020manymodalqa}, \textsc{TaBERT} \cite{yin2020tabert},  NQ-Tables \cite{herzig-etal-2021-open}, FEVEROUS \cite{aly-etal-2021-fact}, FeTaQA \cite{nan-etal-2022-fetaqa}, \textsc{HybriDialogue} \cite{nakamura-etal-2022-hybridialogue}, and HiTab \cite{cheng-etal-2022-hitab}. Other tabular datasets are constructed from financial reports including TAT-QA \cite{zhu-etal-2021-tat}, \textsc{FinQA} \cite{chen-etal-2021-finqa}, \textsc{MultiHiertt} \cite{zhao-etal-2022-multihiertt}, or scientific reviews \cite{sundar2024cpapers}. 

Proposed approaches to address the tabular interpretation task include architectures based off of the Transformer encoder \cite{yin2020tabert,herzig-etal-2020-tapas, chen2019tabfact, eisenschlos-etal-2020-understanding, liu2021tapex, gu-etal-2022-pasta, yang-etal-2022-tableformer}, decoder \cite{gong-etal-2020-tablegpt, akhtar-etal-2023-exploring,zha2023tablegpt,jiang-etal-2023-structgpt,zhang2023tablellama, 10.1145/3616855.3635752,10.1145/3719206}, or both (encoder-decoder) \cite{nakamura-etal-2022-hybridialogue,deng-etal-2022-pacific, sundar-heck-2023-ctbls}. 

\subsection{Tabular Modification}
Tabular modification concerns the manipulation of the content within an existing table without altering the overall structure of rows and columns. 
Early work on tabular modification explored the generation of procedural commands for spreadsheets using synthesis algorithms \cite{singh2012learning, Shigarov2019TabbyXLRS}. Tools utilizing programming-by-example to parse user intents into executable commands have also been explored \cite{scaffidi_intelligently_2009,10.1145/1978942.1979444,10.1145/3035918.3064034, 9953543,9908529,xing2024table}. 
More recent work has shifted focus towards leveraging LLMs to synthesize commands for tools \cite{10304286}, reformat tabular information \cite{dargahi2024dtt}, and execute programming commands \cite{liu-etal-2024-rethinking}. 

\subsection{Tabular Generation}
Tabular generation focuses on expanding an existing table by adding a new row or column.
Research on tabular generation initially employed discriminative techniques, such as tree-based methods for generating tables of contents \cite{branavan-etal-2007-generating} and SVMs to classify text across various labels \citet{aramaki_text2table_2009}. Recent approaches have shifted towards neural techniques including Generative Adversarial Networks (GANs) \cite{xu_synthesizing_2018, park_data_2018, chen_faketables_2019, zhao2021ctab}, Autoencoders \cite{li_evaluating_2019, darabi_synthesising_2021}, Diffusion models \cite{kotelnikov_tabddpm_2023}, and LLMs \cite{borisov_language_2023, solatorio_realtabformer_2023, gulati_tabmt_2023,zhao_docmath-eval_2023, seedat_curated_2024, deng2024texttupletableinformationintegrationtexttotable}. 

A similar line of research also explores the generation of tabular data from associated textual information. \citet{wu_text--table_2022} introduced four datasets and proposed a modification to the Transformer's attention mechanism to summarize textual information in a tabular format by inverting datasets created for the dual task of converting tables to text, (as opposed to new conversational evidence). Other approaches to summarize textual information in a tabular format include the addition of learnable bias parameters \cite{pietruszka_stable_2022} and structure-aware instruction-tuning \cite{tang_struc-bench_2023}. 

In contrast to prior work addressing a single mode of interaction, \mbox{iTBLS} is a dataset unifying tabular interpretation, modification, and generation in a conversational format. Additionally, \mbox{iTBLS} broadens the range of interactions to include mathematical reasoning, natural language manipulation, and the expansion of tables using natural language. Furthermore, by leveraging scientific articles from arXiv as a primary source, \mbox{iTBLS} introduces a novel and rich source of information that is not present in existing datasets. 

\section{The iTBLS Dataset}
\label{sec:dataset_description}
The Interactive Tables (\mbox{iTBLS}) dataset features conversational interactions situated in tabular data, covering the three distinct types of interactions described in Section \ref{sec:related_work}: \textit{interpretation}, \textit{modification}, and \textit{generation}. Each example type is exemplified in Figure \ref{fig:dataset_examples} and described below. In addition, since the mode of interaction is not known a priori, any proposed approach using \mbox{iTBLS} must effectively identify the interaction type, either explicitly or implicitly. In the following sections, we provide a detailed description of each type of interaction and outline the dataset collection process. 

\subsection{Tasks}

{\bf Tabular Interpretation}: In \mbox{iTBLS}, interpretive interactions are structured as question-answer pairs, where the goal is to identify the cell referred to by the question. The references could be absolute (referring to a specific row or column), or relative (referring to one cell in the context of another). Appendix \ref{sec:dataset_examples_itbls} details absolute and relative references in \mbox{iTBLS}. 

\textbf{Tabular modification}: We conceptualize modification in \mbox{iTBLS} as a series of cell swaps, positing that any content rearrangement can ultimately be reduced to such exchanges.  This approach allows for both explicit references, where specific row and column numbers are cited, and implicit references, which rely on the content or relative positions of cells. Table \ref{tab:modif_example} in Appendix \ref{sec:dataset_examples_itbls} showcases examples from \mbox{iTBLS}. As observed, there is a mix of explicit and implicit references to the specific contents to be manipulated. 

\textbf{Tabular generation}: In \mbox{iTBLS}, table generation is guided by new natural language evidence.  This evidence clarifies appending a row or column, defines the suitable header, and supplies the data entries for the new row (or column) relative to existing columns (or rows). This process ensures that the added elements are contextually relevant and accurately integrated into the table. Table \ref{tab:generat_example} in Appendix \ref{sec:dataset_examples_itbls} provides examples of such interactions, demonstrating how users can request the incorporation of new row and column data into an established table framework.

In \mbox{iTBLS}, the mode of interaction is not explicitly stated by the user, introducing an additional task:  \textbf{interaction identification}. This task involves predicting whether the interaction is intended for interpretation, modification, or generation based solely on the user's request.

\subsection{Dataset Collection}
To collect the dataset, first we use \textsc{AxCell} \cite{kardas-etal-2020-axcell} an automatic machine learning pipeline for extracting results from papers. \textsc{AxCell} is used to parse tabular information from papers on arXiv to populate online leaderboards comparing scientific methods. Using \textsc{AxCell}, we collect 20,000 tables from academic papers in Mathematics, Physics, and Computer Science over a period spanning from 2007 to 2014. The tables are processed to remove stray characters resulting from the conversion from \LaTeX. Additionally, only tables with at least three rows and three columns to at most ten rows or ten columns are retained. The final dataset consists of 4000 tables split between train, development, and test sets. 

For each table, we generate three sequential edits corresponding to different types of interaction. Interpretation involves generating a dialogue turn (question-answer pair) grounded on a single cell of the table. Modification involves manipulating two cells of an existing table by swapping them. Finally, generation encompasses the task of appending either a new row or a column to an existing table based on a natural language utterance. 

To enhance the quality of the dataset and minimize errors, we implement a strategic selection process for the table components involved in each interaction. In \textit{interpretation}, a cell is randomly selected to ground the dialogue. For \textit{modification}, two cells are chosen and their positions are swapped to simulate a realistic table manipulation scenario.  In \textit{generation}, all cells in a randomly masked row or column are used as the basis for appending new table data. All of the interactions are based on cells that do not belong to row or column headers, that is, they reside in the body of the table. 

For our dataset creation, we employ two distinct sources for generating dialogue turns based on the type of interaction and the specific table component involved. For tasks related to tabular interpretation and modification, we engage crowd-workers from Amazon Mechanical Turk (AMT). These workers are tasked with formulating questions or commands that pertain to the pre-identified cell(s) designated for each interaction. We recruit workers from Australia, Canada, Ireland, New Zealand, the United Kingdom, and the USA. Each crowdworker is compensated at a rate of \$0.15 per Human Intelligence Task (HIT), with the average completion time for each HIT being approximately 40 seconds. Detailed information on the AMT interface used for these tasks is included in Appendix \ref{sec:app_mturk}.

For generation, \mbox{GPT-4} is prompted to write a dialogue turn summarizing a row or column of the table. 
The prompt is as follows:

\textit{The string contains information from a table [table]. Describe the content in this [row/column] for a visually impaired user in one line. Make sure to include all information from the rows and columns and appropriate headers so the user can understand the content.}

Each sample in the dataset contains the source arXiv ID, the table that the conversation is situated in, the index of that table within the paper (e.g. Table X), the utterance describing the interaction, the ground truth cell(s) involved in the interaction, and finally the expected output.  
Statistics of the datasets are provided in Table \ref{tab:itbls_stats}. 

\begin{table}[h]
    \centering
    \begin{tabular}{l c c c}
    \toprule
        \textbf{Statistic} & \textbf{Interpret} & \textbf{Modify} & \textbf{Generate} \\
    \midrule
        \# Samples & 4168 & 4168 & 4168 \\
        \midrule 
        \multicolumn{4}{l}{\# Per utterance} \\
        Words  & \multirow{1}{*}{10.6 }& \multirow{1}{*}{13.4} & \multirow{1}{*}{31.6} \\
        Tokens & \multirow{1}{*}{14.3}& \multirow{1}{*}{18.3} & \multirow{1}{*}{59.1} \\
        \midrule 
        \multicolumn{4}{l}{\# Per table} \\
        Cells & 28.1 & 28.1 & 25.31 \\
        (Cols/Rows) & 5.0/5.5 & 5.0/5.5 & 4.8/5.3 \\
        \bottomrule
    \end{tabular}
    \caption{Statistics of the iTBLS dataset}
    \label{tab:itbls_stats}
\end{table}

\section{Methods}

\subsection{Table operations through conditional question answering}
We also present a novel approach that reformulates operations on tables as question answering. A primary challenge in tabular operations using LLMs lies in ensuring the syntactic validity of the produced tables. Every row and column in a table must contain the same number of cells, with row and column headers delineating relationships between cells. Failing to adhere to this constraint invalidates the structure of the table and the information presented. Prior work addresses this constraint by including additional parameters like row and column relation embeddings \cite{wu_text--table_2022} or positional bias \cite{pietruszka_stable_2022} to get the model to attend to header cells while generating content. However, this results in highly specialized architectures for a singular task. Breaking the task down into question-answering results in a more interpretable framework while ensuring validity of the generated tables. 

The first step identifies the mode of interaction and the cell(s) the user is referring to, which is used to formulate a question. The second step converts the table into a pandas dataframe, parses the table and the question generated from the previous step to obtain a pandas command corresponding to the task, and executes the command on the dataframe to generate the final table. Generating a valid command ensures that the final table is syntactically valid as well (that is, the number of columns across all rows is consistent). 

For the \textit{interpret} task, the question-answer reformulation is trivial, since all interpretive queries and associated responses are naturally question-answer pairs. For the \textit{modify} task, the question is of the form \textit{To which cells is the user referring?}. A language model is then fine-tuned to generate a response containing the cells (indexed by row and column). Then, the LLM response is reformatted into an appropriate pandas command. Finally, for the \textit{generate} task, the question-answering is more nuanced. First, the user request is parsed to identify whether a row or a column is to be appended. The header of the corresponding row is then extracted from the user request. Using the extracted header and the other header cells of the table, questions are generated for each of the empty cells to be filled in the form \textit{What is the row value for column?}. The user request is parsed to obtain the answers to these generated questions, forming the corresponding row or column to be appended. 

\begin{figure}[t]
    \centering
    \includegraphics[width=\linewidth]{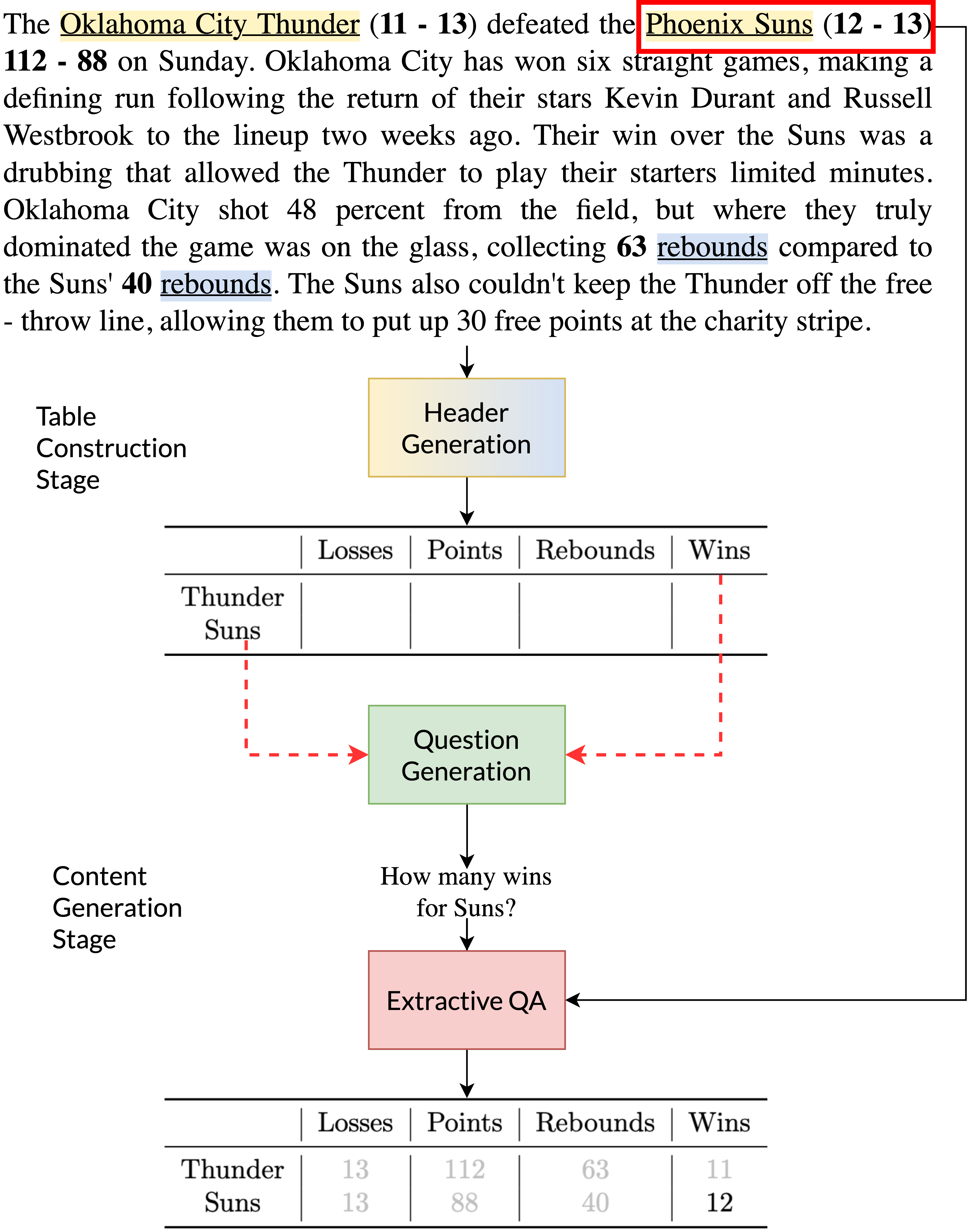}
    \caption{Overview of the novel question-answering reformulation to perform table operations}
    \label{fig:gTBLS}
\end{figure}

\section{Results}

\subsection{Experimental Setup}
For our experiments, we utilize Gemma models \cite{team2024gemma}. We fine-tune the instruction-tuned base model \texttt{gemma-2-9b-it} using LoRA \cite{hu2022lora}. Hyperparameters for our training setup as well as LoRA parameters are shown in Appendix \ref{sec:appendix}.

\subsection{Datasets}
In addition to the iTBLS dataset, we also evaluate our method on five datasets to summarize textual paragraphs to tables \cite{wu_text--table_2022}. While iTBLS is a table-to-table or table-to-text task, the datasets proposed by \citet{wu_text--table_2022} address the dual problem of text-to-table. The datasets consist of textual paragraphs containing some information that is to be converted into a tabular format by determining both the appropriate header cells and the content that the table is filled with.  

\citet{wu_text--table_2022} present datasets for the text-to-table task by inverting datasets created for the dual problem of generating textual descriptions from tables. Each dataset consists of textual paragraphs paired with tabular information summarizing content in the text. Dataset statistics are available in Appendix \ref{sec:Datasetstats}. Each dataset is described below.

E2E \cite{novikova_e2e_2017} concerns restaurant descriptions, requiring summarization of information into tables with descriptors like restaurant name, customer rating, and location.
WikiTableText (WTT) \cite{bao_table--text_2018}, sourced from Wikipedia, consists of natural language descriptions generated from tabular data across various topics. 
WikiBio \cite{lebret-etal-2016-neural} comprises introductions of individuals from Wikipedia alongside tabular summaries extracted from the same page's information box.
In contrast to E2E, the table headers in the WikiTableText and WikiBio datasets vary widely across data samples. 

Example textual paragraphs and associated tables from each dataset are presented in \ref{sec:dataset_examples_t2t}.

\subsection{Metrics}

\textbf{Exact-Match (EM)}: On the iTBLS dataset, we report exact-match, that is, whether or not the generated table matches the ground-truth table exactly. 

\textbf{BERTScore}: On the E2E, WTT, WikiBio and RotoWire datasets, we report BERTScore \cite{zhang_bertscore_2020} in addition to EM to be consistent with prior work. BERTScore is a measure of semantic similarity which computes the similarity of embeddings in a latent space obtained using an encoder language model. 

Consistent with prior work, all our evaluations are order-invariant. That is, credit is given as long as the generated cells are indexed by the correct row and column headers, even if the headers themselves are in different positions between the model-generated response and the ground-truth. 

\subsection{iTBLS}

Results on the iTBLS dataset using a vanilla sequence-to-sequence approach and the question-answering-based method are presented in Table \ref{tab:performance_iTBLS}. As observed in the results, the generate task is the hardest, with performance slightly lower on the generate task when compared to interpret and modify. This is a result of the fact that the exact-match metric only provides credit when all cells are correct (necessitating that all cells in the output are identical to the ground truth) and does not provide partial credit for getting some of the cells right, and the fact that the generate task requires getting more cells right in comparison to the other tasks. 

\begin{table}[h]
    \centering
    \begin{tabular}{l l c}
    \toprule
        \textbf{Split} & \textbf{Approach} & \textbf{Exact-Match} \\
        \midrule 
         \multirow{2}{*}{Interpret} & Seq2seq & 88.29 \\
         & iTBLS as QA & 90.98 \\
         \midrule
         \multirow{2}{*}{Modify} & Seq2seq & 74.65 \\
         & iTBLS as QA & 89.58 \\
         \midrule
         \multirow{2}{*}{Generate} & Seq2seq & 48.94 \\
         & iTBLS as QA & 73.32 \\
         \bottomrule
    \end{tabular}
    \caption{Comparison between the question-answering reformulation and a vanilla sequence-to-sequence modeling approach on the iTBLS dataset}
    \label{tab:performance_iTBLS}
\end{table}

\subsection{Text-to-table}

The results on the text-to-table datasets proposed by \citet{wu_text--table_2022} are available in Table \ref{tab:performance_t2t}. Our method performs on par with or better than the prior state-of-the-art method in terms of BERTScore and is competitive with prior work in terms of Exact-Match. The exact-match score does not reflect true performance on the WikiBio dataset since synonyms are penalized under this framework. A deep-dive into the results is presented in Section \ref{sec:analysis}. 

\begin{table}[h]
    \centering
    \begin{tabular}{l l c c}
    \toprule
       Dataset  & Approach & EM & BS \\
       \midrule 
         \multirow{2}{*}{WTT} & \citet{wu_text--table_2022} & 62.71 & 80.74 \\
         & Ours & 75.96 & 95.52 \\
         \midrule 
         \multirow{2}{*}{Wikibio} & \citet{wu_text--table_2022} & 69.71 & 76.56 \\
         & Ours & 66.65 & 92.60 \\
         \midrule 
         \multirow{2}{*}{E2E} & \citet{wu_text--table_2022} & 97.94 & 98.57 \\
         & Ours & 97.64 & 99.35 \\
         \bottomrule
    \end{tabular}
    \caption{Comparison between our method and prior work on the text to table task in terms of \underline{E}xact-\underline{M}atch and \underline{B}ERT\underline{S}core}
    \label{tab:performance_t2t}
\end{table}

\subsection{Analysis of Errors}
\label{sec:analysis}
An analysis of the difference in performance between the prior state of the art and our approach is presented in Table \ref{tab:difference_answers}. As observed, the dataset is inconsistent in the description of individuals, with no consistent pattern when middle and last names are present. Furthermore, the use of quantifying information in the header as opposed to the table cell results in no credit using the exact-match metric, though the information contained is exactly the same between the prediction and the ground-truth. 
Finally, the datasets often contain textual examples with multiple possible tabular summarizations, all of which are equally valid, further complicating evaluation. In the third example in Table \ref{tab:difference_answers}, the model correctly generates the `Occupation' as a table header while the ground truth contains an erroneous sample, using the phrase `Known for' instead of `Known as'. 

Examples of errors in the iTBLS dataset are provided in Tables \ref{tab:difference_answers_itbls_interpret}, \ref{tab:difference_answers_itbls_modify}, and \ref{tab:difference_answers_itbls_generate}. On the interpret task, the model incorrectly understand the user request, and produces the cell immediately to the right instead of three columns over. On the modify task (Table \ref{tab:difference_answers_itbls_modify}), the model incorrectly understands the references and swaps index (2,3) with (3,2) instead of swapping indices (2,2) and (3,3). On the generate task (Table \ref{tab:difference_answers_itbls_generate}), the model incorrectly places a tuple and hallucinates a value instead of performing the requested action.

\begin{table}[h!]
\centering
\textbf{Text}: What is the value of the cell in row 1 that is three cells to the right of the cell with a value of 12\%? \\
\vspace{0.5em}
{
\resizebox{\columnwidth}{!}{
\begin{tabular}{c|>{\centering\arraybackslash}p{0.11\textwidth}|>{\centering\arraybackslash}p{0.11\textwidth}|>{\centering\arraybackslash}p{0.11\textwidth}|>{\centering\arraybackslash}p{0.11\textwidth}}
\toprule
\toprule
\multicolumn{4}{l}{\textbf{Input Table}:} \\
\midrule
row ID & $\sigma\mu[I0]$ & $\mu[\tau s]$ & $\sigma[\tau s]$ & $\sigma\mu[\tau s]$ \\
\midrule
0 & 13\% & 912.5 $\mu s$ & 91.9 $\mu s$ & 10.1\% \\
1 & 12\% & 18335.7 $\mu s$ & 90.7 $\mu s$ & 10.0\% \\
2 & 12\% & 903.1 $\mu s$ & 1832.7 $\mu s$ & 10.0\% \\
\midrule
\midrule
\multicolumn{4}{l}{\textbf{Ground Truth}: \textcolor{blue}{10.0\%}} \\
\midrule
\multicolumn{4}{l}{\textbf{Prediction}: \textcolor{red}{18335.7 $\mu s$}} \\
\midrule
\bottomrule
\end{tabular}
}
}
\caption{Example error for iTBLS interpret task. Table source: \url{https://arxiv.org/pdf/1411.5458}}
\label{tab:difference_answers_itbls_interpret}
\end{table}

\begin{table}[h!]
    \centering
    \textbf{Text}: Swap row 1 in the second column with row 2 in the third column \\
    \resizebox{\columnwidth}{!}{
    \vspace{0.5em} 
    {
    \begin{tabular}{c|>{\centering\arraybackslash}p{0.1\textwidth}|>{\centering\arraybackslash}p{0.1\textwidth}|>{\centering\arraybackslash}p{0.1\textwidth}}
    \toprule 
    \toprule 
    \multicolumn{4}{l}{\textbf{Input Table}:} \\ 
    \midrule
    row ID & col 1 & col 2 & col 3 \\
    \midrule 
    0 & X & O & X \\
    1 & NaN & O & O \\
    2 & O & X & X \\
    \midrule 
    \midrule 
    \multicolumn{4}{l}{\textbf{Ground Truth}:} \\ 
    \midrule
    row ID & col 1 & col 2 & col 3 \\
    \midrule 
    0 & X & O & X \\
    1 & NaN & \textcolor{blue}{X} & O \\
    2 & O & X & \textcolor{blue}{O} \\
    \midrule 
    \midrule 
    
    \multicolumn{4}{l}{\textbf{Prediction}:} \\ 
    \midrule
    row ID & col 1 & col 2 & col 3 \\
    \midrule 
    0 & X & O & X \\
    1 & NaN & \textcolor{red}{O} & \textcolor{red}{X} \\
    2 & O & \textcolor{red}{O} & \textcolor{red}{X} \\
    \bottomrule
    \end{tabular}
    }
    }
    \caption{Example error for iTBLS modify task. Table source: \url{https://arxiv.org/pdf/1411.4023}}
    \label{tab:difference_answers_itbls_modify}
\end{table}

\begin{table*}[h]
    \centering
    \resizebox{\textwidth}{!}
    {
    \begin{tabular}{p{0.24\textwidth}|p{0.24\textwidth}|p{0.24\textwidth}|p{0.24\textwidth}}
    \toprule 
    \toprule 
    \multicolumn{4}{{C{0.96\textwidth}}}{\textbf{1. Text}: Walter Clarence Henderson (28 February 1891 -- 20 September 1968) was a progressive conservative party member of the Canadian house of commons. He was born in Carberry, Manitoba and became a farmer by career. He was elected at the Cariboo riding in the 1958 general election, defeating social credit incumbent Bert Leboe.} \\ 
    \multicolumn{4}{{C{0.96\textwidth}}}{\textbf{Generated Table}:} \\ 
    \midrule
    Predicted Header & Prediction - iTBLS & Ground Truth Header &  Ground Truth \\
    \midrule 
    Name & Walter Henderson & Name & Walter \textcolor{OliveGreen}{Clarence} Henderson \\
    Profession & Farmer & Profession & Farmer \\
    Party & Progressive Conservative & Party &  Progressive Conservative \\
    \midrule 
    \midrule 
    \multicolumn{4}{{C{0.96\textwidth}}}{\textbf{2. Text}: The production of Tautona mine is 235,000 ounces in 2013.} \\ 
    \multicolumn{4}{{C{0.96\textwidth}}}{\textbf{Generated Table}:} \\ 
    \midrule
    Predicted Header & Prediction - iTBLS & Ground Truth Header &  Ground Truth \\
    \midrule 
    Title & Tautona mine & Title & Tautona mine \\
    Subtitle & Production & Subtitle & Production \\
    Year & 2013 & Year & 2013 \\
    Production \textcolor{red}{(ounces)} & 235,000 & Production & 235,00 \textcolor{OliveGreen}{ounces} \\
    \midrule 
    \midrule 
    \multicolumn{4}{{C{0.96\textwidth}}}{\textbf{3. Text}: Elango Nagarajah, also known as ``Thaimann Elango", is a Tamil film actor, director, producer and lyricist in the Tamil film industry. He began his career in his early ages as a producer for the Tamil film Anbudan, starred Arun Vijay, Meena, Rambha (actress) in the main was released in the year 2000.} \\ 
    \multicolumn{4}{{C{0.96\textwidth}}}{\textbf{Generated Table}:} \\ 
    \midrule
    Predicted Header & Prediction - iTBLS & Ground Truth Header &  Ground Truth \\
    \midrule 
    Name & Elango \textcolor{red}{Nagarajah} & Name & Elango \\
    Occupation & actor, director, producer, lyricist & Known for & Thaimann \\
    \bottomrule
    \bottomrule
    \end{tabular}
    }
    \caption{Difference between the tables generated by the Zero Shot (ZS) and Fine-Tuned (FT) approaches with respect to the Ground Truth on the WikiBio and WikiTableText datasets with additions and deletions represented using \textcolor{red}{red} and \textcolor{OliveGreen}{green}.}
    \label{tab:difference_answers}
\end{table*}

\section{Conclusion}
This paper introduces Interactive Tables (iTBLS), a dataset of interactive conversations addressing three types of tasks -- interpretation, modification, and generation. In contrast to prior tabular datasets that are sourced from Wikipedia or financial reports, iTBLS is situated in tabular data obtained from scientific pre-prints on ArXiv. Success on the iTBLS dataset requires understanding both ordinal and cardinal references to cell positions, and understanding implicit references. Additionally, the paper introduces a novel framework that reformulates tabular operations as question-answering. Appropriate questions are created based on the input table and the nature of interaction, and the user request is used as evidence to obtain the answers. The developed approach demonstrates an improvement over a sequence-to-sequence modeling approach on the iTBLS dataset. In addition, the question-answering-based reformulation is evaluated on datasets for the text-to-table task, obtaining up to 13\% improvement in terms of exact-match accuracy and 16\% improvement in terms of BERTScore compared to the prior state-of-the-art. 

\section*{Limitations}

While iTBLS introduces a dataset for interactive conversations over tabular information, there are some avenues for improvement. In this dataset, modification is modeled as a series of swaps. A more comprehensive sequence of manipulations includes in-place modification of values and modifying a cell’s value based on other cells using both absolute and relative references. While sourcing tabular information from arXiv provides a cost-efficient approach, LLMs are often pre-trained on \LaTeX sources from arXiv. This paper alleviates the issue by sourcing natural language commands from crowdworkers. Future work could look at collecting tabular information from crowdworkers as well. While we present a suite of baseline approaches for iTBLS, there is still headroom between the presented approaches and perfect performance. We identify the closure of this gap as an avenue for
future work.

\section*{Acknowledgments}
This work was supported by CoCoSys, one of seven centers in JUMP 2.0, a Semiconductor Research Corporation (SRC) program sponsored by DARPA.

\bibliography{acl_latex}

\begin{thebibliography}{66}
\providecommand{\natexlab}[1]{#1}

\bibitem[{Akhtar et~al.(2023)Akhtar, Shankarampeta, Gupta, Patil, Cocarascu,
  and Simperl}]{akhtar-etal-2023-exploring}
Mubashara Akhtar, Abhilash Shankarampeta, Vivek Gupta, Arpit Patil, Oana
  Cocarascu, and Elena Simperl. 2023.
\newblock \href {https://doi.org/10.18653/v1/2023.findings-emnlp.1028}
  {Exploring the numerical reasoning capabilities of language models: A
  comprehensive analysis on tabular data}.
\newblock In \emph{Findings of the Association for Computational Linguistics:
  EMNLP 2023}, pages 15391--15405, Singapore. Association for Computational
  Linguistics.

\bibitem[{Aly et~al.(2021)Aly, Guo, Schlichtkrull, Thorne, Vlachos,
  Christodoulopoulos, Cocarascu, and Mittal}]{aly-etal-2021-fact}
Rami Aly, Zhijiang Guo, Michael~Sejr Schlichtkrull, James Thorne, Andreas
  Vlachos, Christos Christodoulopoulos, Oana Cocarascu, and Arpit Mittal. 2021.
\newblock \href {https://doi.org/10.18653/v1/2021.fever-1.1} {The fact
  extraction and {VER}ification over unstructured and structured information
  ({FEVEROUS}) shared task}.
\newblock In \emph{Proceedings of the Fourth Workshop on Fact Extraction and
  VERification (FEVER)}, pages 1--13, Dominican Republic. Association for
  Computational Linguistics.

\bibitem[{Aramaki et~al.(2009)Aramaki, Miura, Tonoike, Ohkuma, Mashuichi, and
  Ohe}]{aramaki_text2table_2009}
Eiji Aramaki, Yasuhide Miura, Masatsugu Tonoike, Tomoko Ohkuma, Hiroshi
  Mashuichi, and Kazuhiko Ohe. 2009.
\newblock \href {https://doi.org/10.3115/1572364.1572390} {{TEXT2TABLE}:
  medical text summarization system based on named entity recognition and
  modality identification}.
\newblock In \emph{Proceedings of the {Workshop} on {BioNLP} - {BioNLP} '09},
  page 185, Boulder, Colorado. Association for Computational Linguistics.

\bibitem[{Bao et~al.(2018)Bao, Tang, Duan, Yan, Lv, Zhou, and
  Zhao}]{bao_table--text_2018}
Junwei Bao, Duyu Tang, Nan Duan, Zhao Yan, Yuanhua Lv, Ming Zhou, and Tiejun
  Zhao. 2018.
\newblock \href {http://arxiv.org/abs/1805.11234} {Table-to-{Text}:
  {Describing} {Table} {Region} with {Natural} {Language}}.
\newblock \emph{arXiv preprint}.
\newblock ArXiv:1805.11234 [cs].

\bibitem[{Borisov et~al.(2023)Borisov, Sessler, Leemann, Pawelczyk, and
  Kasneci}]{borisov_language_2023}
Vadim Borisov, Kathrin Sessler, Tobias Leemann, Martin Pawelczyk, and Gjergji
  Kasneci. 2023.
\newblock \href {https://openreview.net/forum?id=cEygmQNOeI} {Language {Models}
  are {Realistic} {Tabular} {Data} {Generators}}.
\newblock In \emph{The {Eleventh} {International} {Conference} on {Learning}
  {Representations}}.

\bibitem[{Branavan et~al.(2007)Branavan, Deshpande, and
  Barzilay}]{branavan-etal-2007-generating}
S.~R.~K. Branavan, Pawan Deshpande, and Regina Barzilay. 2007.
\newblock \href {https://aclanthology.org/P07-1069/} {Generating a
  table-of-contents}.
\newblock In \emph{Proceedings of the 45th Annual Meeting of the Association of
  Computational Linguistics}, pages 544--551, Prague, Czech Republic.
  Association for Computational Linguistics.

\bibitem[{Chen et~al.(2019{\natexlab{a}})Chen, Jajodia, Liu, Park, Sokolov, and
  Subrahmanian}]{chen_faketables_2019}
Haipeng Chen, Sushil Jajodia, Jing Liu, Noseong Park, Vadim Sokolov, and V.~S.
  Subrahmanian. 2019{\natexlab{a}}.
\newblock \href {https://doi.org/10.24963/ijcai.2019/287} {{FakeTables}:
  {Using} {GANs} to {Generate} {Functional} {Dependency} {Preserving} {Tables}
  with {Bounded} {Real} {Data}}.
\newblock In \emph{Proceedings of the {Twenty}-{Eighth} {International} {Joint}
  {Conference} on {Artificial} {Intelligence}}, pages 2074--2080, Macao, China.
  International Joint Conferences on Artificial Intelligence Organization.

\bibitem[{Chen et~al.(2023)Chen, Weng, Huang, Shu, Zhou, Sun, and Wu}]{9908529}
Ran Chen, Di~Weng, Yanwei Huang, Xinhuan Shu, Jiayi Zhou, Guodao Sun, and
  Yingcai Wu. 2023.
\newblock \href {https://doi.org/10.1109/TVCG.2022.3209385} {Rigel:
  Transforming tabular data by declarative mapping}.
\newblock \emph{IEEE Transactions on Visualization and Computer Graphics},
  29(1):128--138.

\bibitem[{Chen et~al.(2019{\natexlab{b}})Chen, Wang, Chen, Zhang, Wang, Li,
  Zhou, and Wang}]{chen2019tabfact}
Wenhu Chen, Hongmin Wang, Jianshu Chen, Yunkai Zhang, Hong Wang, Shiyang Li,
  Xiyou Zhou, and William~Yang Wang. 2019{\natexlab{b}}.
\newblock Tabfact: A large-scale dataset for table-based fact verification.
\newblock \emph{arXiv preprint arXiv:1909.02164}.

\bibitem[{Chen et~al.(2021)Chen, Chen, Smiley, Shah, Borova, Langdon, Moussa,
  Beane, Huang, Routledge, and Wang}]{chen-etal-2021-finqa}
Zhiyu Chen, Wenhu Chen, Charese Smiley, Sameena Shah, Iana Borova, Dylan
  Langdon, Reema Moussa, Matt Beane, Ting-Hao Huang, Bryan Routledge, and
  William~Yang Wang. 2021.
\newblock \href {https://doi.org/10.18653/v1/2021.emnlp-main.300} {{F}in{QA}: A
  dataset of numerical reasoning over financial data}.
\newblock In \emph{Proceedings of the 2021 Conference on Empirical Methods in
  Natural Language Processing}, pages 3697--3711, Online and Punta Cana,
  Dominican Republic. Association for Computational Linguistics.

\bibitem[{Cheng et~al.(2022)Cheng, Dong, Wang, Jia, Guo, Gao, Han, Lou, and
  Zhang}]{cheng-etal-2022-hitab}
Zhoujun Cheng, Haoyu Dong, Zhiruo Wang, Ran Jia, Jiaqi Guo, Yan Gao, Shi Han,
  Jian-Guang Lou, and Dongmei Zhang. 2022.
\newblock \href {https://doi.org/10.18653/v1/2022.acl-long.78} {{H}i{T}ab: A
  hierarchical table dataset for question answering and natural language
  generation}.
\newblock In \emph{Proceedings of the 60th Annual Meeting of the Association
  for Computational Linguistics (Volume 1: Long Papers)}, pages 1094--1110,
  Dublin, Ireland. Association for Computational Linguistics.

\bibitem[{Cremaschi et~al.(2025)Cremaschi, D'Adda, and
  Maurino}]{10.1145/3719206}
Marco Cremaschi, Fabio D'Adda, and Andrea Maurino. 2025.
\newblock \href {https://doi.org/10.1145/3719206} {steellm: An llm for
  generating semantic annotations of tabular data}.
\newblock \emph{ACM Trans. Intell. Syst. Technol.}
\newblock Just Accepted.

\bibitem[{Darabi and Elor(2021)}]{darabi_synthesising_2021}
Sajad Darabi and Yotam Elor. 2021.
\newblock Synthesising multi-modal minority samples for tabular data.
\newblock \emph{arXiv preprint arXiv:2105.08204}.

\bibitem[{Dargahi~Nobari and Rafiei(2024)}]{dargahi2024dtt}
Arash Dargahi~Nobari and Davood Rafiei. 2024.
\newblock Dtt: An example-driven tabular transformer for joinability by
  leveraging large language models.
\newblock \emph{Proceedings of the ACM on Management of Data}, 2(1):1--24.

\bibitem[{Deng et~al.(2022)Deng, Lei, Zhang, Lam, and
  Chua}]{deng-etal-2022-pacific}
Yang Deng, Wenqiang Lei, Wenxuan Zhang, Wai Lam, and Tat-Seng Chua. 2022.
\newblock \href {https://doi.org/10.18653/v1/2022.emnlp-main.469} {{PACIFIC}:
  Towards proactive conversational question answering over tabular and textual
  data in finance}.
\newblock In \emph{Proceedings of the 2022 Conference on Empirical Methods in
  Natural Language Processing}, pages 6970--6984, Abu Dhabi, United Arab
  Emirates. Association for Computational Linguistics.

\bibitem[{Deng et~al.(2024)Deng, Chan, Wang, Sun, Fan, Zheng, Yim, and
  Song}]{deng2024texttupletableinformationintegrationtexttotable}
Zheye Deng, Chunkit Chan, Weiqi Wang, Yuxi Sun, Wei Fan, Tianshi Zheng, Yauwai
  Yim, and Yangqiu Song. 2024.
\newblock \href {https://arxiv.org/abs/2404.14215} {Text-tuple-table: Towards
  information integration in text-to-table generation via global tuple
  extraction}.
\newblock \emph{Preprint}, arXiv:2404.14215.

\bibitem[{Eisenschlos et~al.(2020)Eisenschlos, Krichene, and
  M{\"u}ller}]{eisenschlos-etal-2020-understanding}
Julian Eisenschlos, Syrine Krichene, and Thomas M{\"u}ller. 2020.
\newblock \href {https://doi.org/10.18653/v1/2020.findings-emnlp.27}
  {Understanding tables with intermediate pre-training}.
\newblock In \emph{Findings of the Association for Computational Linguistics:
  EMNLP 2020}, pages 281--296, Online. Association for Computational
  Linguistics.

\bibitem[{Fang et~al.(2024)Fang, Xu, Anting~Tan, Zhang, Hu, Qi, Nickleach,
  Socolinsky, Sengamedu, and Faloutsos}]{fang2024large}
Xi~Fang, Weijie Xu, Fiona Anting~Tan, Jiani Zhang, Ziqing Hu, Yanjun Qi, Scott
  Nickleach, Diego Socolinsky, Srinivasan Sengamedu, and Christos Faloutsos.
  2024.
\newblock Large language models on tabular data--a survey.
\newblock \emph{arXiv e-prints}, pages arXiv--2402.

\bibitem[{Gong et~al.(2020)Gong, Sun, Feng, Qin, Bi, Liu, and
  Liu}]{gong-etal-2020-tablegpt}
Heng Gong, Yawei Sun, Xiaocheng Feng, Bing Qin, Wei Bi, Xiaojiang Liu, and Ting
  Liu. 2020.
\newblock \href {https://doi.org/10.18653/v1/2020.coling-main.179}
  {{T}able{GPT}: Few-shot table-to-text generation with table structure
  reconstruction and content matching}.
\newblock In \emph{Proceedings of the 28th International Conference on
  Computational Linguistics}, pages 1978--1988, Barcelona, Spain (Online).
  International Committee on Computational Linguistics.

\bibitem[{Gu et~al.(2022)Gu, Fan, Tang, Nakov, Zhao, and
  Du}]{gu-etal-2022-pasta}
Zihui Gu, Ju~Fan, Nan Tang, Preslav Nakov, Xiaoman Zhao, and Xiaoyong Du. 2022.
\newblock \href {https://doi.org/10.18653/v1/2022.emnlp-main.331} {{PASTA}:
  Table-operations aware fact verification via sentence-table cloze
  pre-training}.
\newblock In \emph{Proceedings of the 2022 Conference on Empirical Methods in
  Natural Language Processing}, pages 4971--4983, Abu Dhabi, United Arab
  Emirates. Association for Computational Linguistics.

\bibitem[{Gulati and Roysdon(2023)}]{gulati_tabmt_2023}
Manbir~S. Gulati and Paul~F. Roysdon. 2023.
\newblock \href {https://openreview.net/forum?id=qs4swxtIAQ} {{TabMT}:
  {Generating} tabular data with masked transformers}.
\newblock In \emph{Thirty-seventh {Conference} on {Neural} {Information}
  {Processing} {Systems}}.

\bibitem[{Hannan et~al.(2020)Hannan, Jain, and Bansal}]{hannan2020manymodalqa}
Darryl Hannan, Akshay Jain, and Mohit Bansal. 2020.
\newblock Manymodalqa: Modality disambiguation and qa over diverse inputs.
\newblock In \emph{Proceedings of the AAAI conference on artificial
  intelligence}, volume~34, pages 7879--7886.

\bibitem[{Herzig et~al.(2021)Herzig, M{\"u}ller, Krichene, and
  Eisenschlos}]{herzig-etal-2021-open}
Jonathan Herzig, Thomas M{\"u}ller, Syrine Krichene, and Julian Eisenschlos.
  2021.
\newblock \href {https://doi.org/10.18653/v1/2021.naacl-main.43} {Open domain
  question answering over tables via dense retrieval}.
\newblock In \emph{Proceedings of the 2021 Conference of the North American
  Chapter of the Association for Computational Linguistics: Human Language
  Technologies}, pages 512--519, Online. Association for Computational
  Linguistics.

\bibitem[{Herzig et~al.(2020)Herzig, Nowak, M{\"u}ller, Piccinno, and
  Eisenschlos}]{herzig-etal-2020-tapas}
Jonathan Herzig, Pawel~Krzysztof Nowak, Thomas M{\"u}ller, Francesco Piccinno,
  and Julian Eisenschlos. 2020.
\newblock \href {https://doi.org/10.18653/v1/2020.acl-main.398} {{T}a{P}as:
  Weakly supervised table parsing via pre-training}.
\newblock In \emph{Proceedings of the 58th Annual Meeting of the Association
  for Computational Linguistics}, pages 4320--4333, Online. Association for
  Computational Linguistics.

\bibitem[{Hu et~al.(2022)Hu, Shen, Wallis, Allen-Zhu, Li, Wang, Wang, Chen
  et~al.}]{hu2022lora}
Edward~J Hu, Yelong Shen, Phillip Wallis, Zeyuan Allen-Zhu, Yuanzhi Li, Shean
  Wang, Lu~Wang, Weizhu Chen, et~al. 2022.
\newblock Lora: Low-rank adaptation of large language models.
\newblock \emph{ICLR}, 1(2):3.

\bibitem[{Huang et~al.(2024)Huang, Zhou, Chen, Pan, Shu, Weng, and
  Wu}]{10304286}
Yanwei Huang, Yunfan Zhou, Ran Chen, Changhao Pan, Xinhuan Shu, Di~Weng, and
  Yingcai Wu. 2024.
\newblock \href {https://doi.org/10.1109/TVCG.2023.3329120} {Interactive table
  synthesis with natural language}.
\newblock \emph{IEEE Transactions on Visualization and Computer Graphics},
  30(9):6130--6145.

\bibitem[{Jiang et~al.(2023)Jiang, Zhou, Dong, Ye, Zhao, and
  Wen}]{jiang-etal-2023-structgpt}
Jinhao Jiang, Kun Zhou, Zican Dong, Keming Ye, Xin Zhao, and Ji-Rong Wen. 2023.
\newblock \href {https://doi.org/10.18653/v1/2023.emnlp-main.574}
  {{S}truct{GPT}: A general framework for large language model to reason over
  structured data}.
\newblock In \emph{Proceedings of the 2023 Conference on Empirical Methods in
  Natural Language Processing}, pages 9237--9251, Singapore. Association for
  Computational Linguistics.

\bibitem[{Jin et~al.(2017)Jin, Anderson, Cafarella, and
  Jagadish}]{10.1145/3035918.3064034}
Zhongjun Jin, Michael~R. Anderson, Michael Cafarella, and H.~V. Jagadish. 2017.
\newblock \href {https://doi.org/10.1145/3035918.3064034} {Foofah: Transforming
  data by example}.
\newblock In \emph{Proceedings of the 2017 ACM International Conference on
  Management of Data}, SIGMOD '17, page 683–698, New York, NY, USA.
  Association for Computing Machinery.

\bibitem[{Kandel et~al.(2011)Kandel, Paepcke, Hellerstein, and
  Heer}]{10.1145/1978942.1979444}
Sean Kandel, Andreas Paepcke, Joseph Hellerstein, and Jeffrey Heer. 2011.
\newblock \href {https://doi.org/10.1145/1978942.1979444} {Wrangler:
  interactive visual specification of data transformation scripts}.
\newblock In \emph{Proceedings of the SIGCHI Conference on Human Factors in
  Computing Systems}, CHI '11, page 3363–3372, New York, NY, USA. Association
  for Computing Machinery.

\bibitem[{Kardas et~al.(2020)Kardas, Czapla, Stenetorp, Ruder, Riedel, Taylor,
  and Stojnic}]{kardas-etal-2020-axcell}
Marcin Kardas, Piotr Czapla, Pontus Stenetorp, Sebastian Ruder, Sebastian
  Riedel, Ross Taylor, and Robert Stojnic. 2020.
\newblock \href {https://doi.org/10.18653/v1/2020.emnlp-main.692} {{AxCell}:
  Automatic extraction of results from machine learning papers}.
\newblock In \emph{Proceedings of the 2020 Conference on Empirical Methods in
  Natural Language Processing (EMNLP)}, pages 8580--8594, Online. Association
  for Computational Linguistics.

\bibitem[{Kotelnikov et~al.(2023)Kotelnikov, Baranchuk, Rubachev, and
  Babenko}]{kotelnikov_tabddpm_2023}
Akim Kotelnikov, Dmitry Baranchuk, Ivan Rubachev, and Artem Babenko. 2023.
\newblock {TabDDPM}: modelling tabular data with diffusion models.
\newblock In \emph{Proceedings of the 40th {International} {Conference} on
  {Machine} {Learning}}, pages 17564--17579.

\bibitem[{Lebret et~al.(2016)Lebret, Grangier, and
  Auli}]{lebret-etal-2016-neural}
R{\'e}mi Lebret, David Grangier, and Michael Auli. 2016.
\newblock \href {https://doi.org/10.18653/v1/D16-1128} {Neural text generation
  from structured data with application to the biography domain}.
\newblock In \emph{Proceedings of the 2016 Conference on Empirical Methods in
  Natural Language Processing}, pages 1203--1213, Austin, Texas. Association
  for Computational Linguistics.

\bibitem[{Li et~al.(2019)Li, Tai, and Huang}]{li_evaluating_2019}
Szu-Chuang Li, Bo-Chen Tai, and Yennun Huang. 2019.
\newblock \href {https://doi.org/10.1109/PRDC47002.2019.00050} {Evaluating
  {Variational} {Autoencoder} as a {Private} {Data} {Release} {Mechanism} for
  {Tabular} {Data}}.
\newblock In \emph{2019 {IEEE} 24th {Pacific} {Rim} {International} {Symposium}
  on {Dependable} {Computing} ({PRDC})}, pages 198--1988.

\bibitem[{Liu et~al.(2021)Liu, Chen, Guo, Ziyadi, Lin, Chen, and
  Lou}]{liu2021tapex}
Qian Liu, Bei Chen, Jiaqi Guo, Morteza Ziyadi, Zeqi Lin, Weizhu Chen, and
  Jian-Guang Lou. 2021.
\newblock Tapex: Table pre-training via learning a neural sql executor.
\newblock \emph{arXiv preprint arXiv:2107.07653}.

\bibitem[{Liu et~al.(2024)Liu, Wang, and Chen}]{liu-etal-2024-rethinking}
Tianyang Liu, Fei Wang, and Muhao Chen. 2024.
\newblock \href {https://doi.org/10.18653/v1/2024.naacl-long.26} {Rethinking
  tabular data understanding with large language models}.
\newblock In \emph{Proceedings of the 2024 Conference of the North American
  Chapter of the Association for Computational Linguistics: Human Language
  Technologies (Volume 1: Long Papers)}, pages 450--482, Mexico City, Mexico.
  Association for Computational Linguistics.

\bibitem[{Nakamura et~al.(2022{\natexlab{a}})Nakamura, Levy, Tuan, Chen, and
  Wang}]{nakamura_hybridialogue_2022}
Kai Nakamura, Sharon Levy, Yi-Lin Tuan, Wenhu Chen, and William~Yang Wang.
  2022{\natexlab{a}}.
\newblock \href {https://doi.org/10.18653/v1/2022.findings-acl.41}
  {{H}ybri{D}ialogue: An information-seeking dialogue dataset grounded on
  tabular and textual data}.
\newblock In \emph{Findings of the Association for Computational Linguistics:
  ACL 2022}, pages 481--492, Dublin, Ireland. Association for Computational
  Linguistics.

\bibitem[{Nakamura et~al.(2022{\natexlab{b}})Nakamura, Levy, Tuan, Chen, and
  Wang}]{nakamura-etal-2022-hybridialogue}
Kai Nakamura, Sharon Levy, Yi-Lin Tuan, Wenhu Chen, and William~Yang Wang.
  2022{\natexlab{b}}.
\newblock \href {https://doi.org/10.18653/v1/2022.findings-acl.41}
  {{H}ybri{D}ialogue: An information-seeking dialogue dataset grounded on
  tabular and textual data}.
\newblock In \emph{Findings of the Association for Computational Linguistics:
  ACL 2022}, pages 481--492, Dublin, Ireland. Association for Computational
  Linguistics.

\bibitem[{Nan et~al.(2022)Nan, Hsieh, Mao, Lin, Verma, Zhang,
  Kry{\'s}ci{\'n}ski, Schoelkopf, Kong, Tang, Mutuma, Rosand, Trindade,
  Bandaru, Cunningham, Xiong, Radev, and Radev}]{nan-etal-2022-fetaqa}
Linyong Nan, Chiachun Hsieh, Ziming Mao, Xi~Victoria Lin, Neha Verma, Rui
  Zhang, Wojciech Kry{\'s}ci{\'n}ski, Hailey Schoelkopf, Riley Kong, Xiangru
  Tang, Mutethia Mutuma, Ben Rosand, Isabel Trindade, Renusree Bandaru, Jacob
  Cunningham, Caiming Xiong, Dragomir Radev, and Dragomir Radev. 2022.
\newblock \href {https://doi.org/10.1162/tacl_a_00446} {{F}e{T}a{QA}: Free-form
  table question answering}.
\newblock \emph{Transactions of the Association for Computational Linguistics},
  10:35--49.

\bibitem[{Novikova et~al.(2017)Novikova, Dušek, and
  Rieser}]{novikova_e2e_2017}
Jekaterina Novikova, Ondřej Dušek, and Verena Rieser. 2017.
\newblock \href {http://arxiv.org/abs/1706.09254} {The {E2E} {Dataset}: {New}
  {Challenges} {For} {End}-to-{End} {Generation}}.
\newblock \emph{arXiv preprint}.
\newblock ArXiv:1706.09254 [cs].

\bibitem[{Park et~al.(2018)Park, Mohammadi, Gorde, Jajodia, Park, and
  Kim}]{park_data_2018}
Noseong Park, Mahmoud Mohammadi, Kshitij Gorde, Sushil Jajodia, Hongkyu Park,
  and Youngmin Kim. 2018.
\newblock \href {https://doi.org/10.14778/3231751.3231757} {Data synthesis
  based on generative adversarial networks}.
\newblock \emph{Proceedings of the VLDB Endowment}, 11(10):1071--1083.

\bibitem[{Pasupat and Liang(2015)}]{pasupat-liang-2015-compositional}
Panupong Pasupat and Percy Liang. 2015.
\newblock \href {https://doi.org/10.3115/v1/P15-1142} {Compositional semantic
  parsing on semi-structured tables}.
\newblock In \emph{Proceedings of the 53rd Annual Meeting of the Association
  for Computational Linguistics and the 7th International Joint Conference on
  Natural Language Processing (Volume 1: Long Papers)}, pages 1470--1480,
  Beijing, China. Association for Computational Linguistics.

\bibitem[{Petricek et~al.(2023)Petricek, Burg, Nazábal, Ceritli,
  Jiménez-Ruiz, and Williams}]{9953543}
Tomas Petricek, Gerrit J. J. van~den Burg, Alfredo Nazábal, Taha Ceritli,
  Ernesto Jiménez-Ruiz, and Christopher K.~I. Williams. 2023.
\newblock \href {https://doi.org/10.1109/TKDE.2022.3222538} {Ai assistants: A
  framework for semi-automated data wrangling}.
\newblock \emph{IEEE Transactions on Knowledge and Data Engineering},
  35(9):9295--9306.

\bibitem[{Pietruszka et~al.(2022)Pietruszka, Turski, Borchmann, Dwojak, Pałka,
  Szyndler, Jurkiewicz, and Garncarek}]{pietruszka_stable_2022}
Michał Pietruszka, Michał Turski, Łukasz Borchmann, Tomasz Dwojak, Gabriela
  Pałka, Karolina Szyndler, Dawid Jurkiewicz, and Łukasz Garncarek. 2022.
\newblock \href {http://arxiv.org/abs/2206.04045} {{STable}: {Table}
  {Generation} {Framework} for {Encoder}-{Decoder} {Models}}.
\newblock \emph{arXiv preprint}.
\newblock ArXiv:2206.04045 [cs].

\bibitem[{Scaffidi et~al.(2009)Scaffidi, Myers, and
  Shaw}]{scaffidi_intelligently_2009}
Christopher Scaffidi, Brad Myers, and Mary Shaw. 2009.
\newblock \href {https://doi.org/10.1145/1502650.1502692} {Intelligently
  creating and recommending reusable reformatting rules}.
\newblock In \emph{Proceedings of the 14th International Conference on
  Intelligent User Interfaces}, IUI '09, page 297–306, New York, NY, USA.
  Association for Computing Machinery.

\bibitem[{Seedat et~al.(2024)Seedat, Huynh, Breugel, and
  Schaar}]{seedat_curated_2024}
Nabeel Seedat, Nicolas Huynh, Boris~van Breugel, and Mihaela van~der Schaar.
  2024.
\newblock Curated {LLM}: {Synergy} of {LLMs} and {Data} {Curation} for tabular
  augmentation in ultra low-data regimes.
\newblock \_eprint: 2312.12112.

\bibitem[{Shigarov et~al.(2019)Shigarov, Khristyuk, Mikhailov, and
  Paramonov}]{Shigarov2019TabbyXLRS}
Alexey~O. Shigarov, Vasiliy~V. Khristyuk, Andrey~A. Mikhailov, and
  Viacheslav~V. Paramonov. 2019.
\newblock \href {https://api.semanticscholar.org/CorpusID:203658267} {Tabbyxl:
  Rule-based spreadsheet data extraction and transformation}.
\newblock In \emph{International Conference on Information and Software
  Technologies}.

\bibitem[{Singh and Gulwani(2012)}]{singh2012learning}
Rishabh Singh and Sumit Gulwani. 2012.
\newblock Learning semantic string transformations from examples.
\newblock \emph{arXiv preprint arXiv:1204.6079}.

\bibitem[{Solatorio and Dupriez(2023)}]{solatorio_realtabformer_2023}
Aivin~V. Solatorio and Olivier Dupriez. 2023.
\newblock \href {http://arxiv.org/abs/2302.02041} {{REaLTabFormer}:
  {Generating} {Realistic} {Relational} and {Tabular} {Data} using
  {Transformers}}.
\newblock \emph{arXiv preprint}.
\newblock ArXiv:2302.02041 [cs].

\bibitem[{Sui et~al.(2024)Sui, Zhou, Zhou, Han, and
  Zhang}]{10.1145/3616855.3635752}
Yuan Sui, Mengyu Zhou, Mingjie Zhou, Shi Han, and Dongmei Zhang. 2024.
\newblock \href {https://doi.org/10.1145/3616855.3635752} {Table meets llm: Can
  large language models understand structured table data? a benchmark and
  empirical study}.
\newblock In \emph{Proceedings of the 17th ACM International Conference on Web
  Search and Data Mining}, WSDM '24, page 645–654, New York, NY, USA.
  Association for Computing Machinery.

\bibitem[{Sundar and Heck(2022)}]{sundar_multimodal_2022}
Anirudh Sundar and Larry Heck. 2022.
\newblock \href {https://doi.org/10.18653/v1/2022.nlp4convai-1.12} {Multimodal
  conversational {AI}: A survey of datasets and approaches}.
\newblock In \emph{Proceedings of the 4th Workshop on NLP for Conversational
  AI}, pages 131--147, Dublin, Ireland. Association for Computational
  Linguistics.

\bibitem[{Sundar et~al.(2024)Sundar, Xu, Gay, Richardson, and
  Heck}]{sundar2024cpapers}
Anirudh Sundar, Jin Xu, William Gay, Christopher Richardson, and Larry Heck.
  2024.
\newblock cpapers: A dataset of situated and multimodal interactive
  conversations in scientific papers.
\newblock \emph{Advances in Neural Information Processing Systems},
  37:66283--66304.

\bibitem[{Sundar and Heck(2023)}]{sundar-heck-2023-ctbls}
Anirudh~S. Sundar and Larry Heck. 2023.
\newblock \href {https://doi.org/10.18653/v1/2023.nlp4convai-1.6} {c{TBLS}:
  Augmenting large language models with conversational tables}.
\newblock In \emph{Proceedings of the 5th Workshop on NLP for Conversational AI
  (NLP4ConvAI 2023)}, pages 59--70, Toronto, Canada. Association for
  Computational Linguistics.

\bibitem[{Tang et~al.(2023)Tang, Zong, Phang, Zhao, Zhou, Cohan, and
  Gerstein}]{tang_struc-bench_2023}
Xiangru Tang, Yiming Zong, Jason Phang, Yilun Zhao, Wangchunshu Zhou, Arman
  Cohan, and Mark Gerstein. 2023.
\newblock \href {http://arxiv.org/abs/2309.08963} {Struc-{Bench}: {Are} {Large}
  {Language} {Models} {Really} {Good} at {Generating} {Complex} {Structured}
  {Data}?}
\newblock \emph{arXiv preprint}.
\newblock ArXiv:2309.08963 [cs].

\bibitem[{Team et~al.(2024)Team, Mesnard, Hardin, Dadashi, Bhupatiraju, Pathak,
  Sifre, Rivi{\`e}re, Kale, Love et~al.}]{team2024gemma}
Gemma Team, Thomas Mesnard, Cassidy Hardin, Robert Dadashi, Surya Bhupatiraju,
  Shreya Pathak, Laurent Sifre, Morgane Rivi{\`e}re, Mihir~Sanjay Kale,
  Juliette Love, et~al. 2024.
\newblock Gemma: Open models based on gemini research and technology.
\newblock \emph{arXiv preprint arXiv:2403.08295}.

\bibitem[{Wu et~al.(2022)Wu, Zhang, and Li}]{wu_text--table_2022}
Xueqing Wu, Jiacheng Zhang, and Hang Li. 2022.
\newblock \href {http://arxiv.org/abs/2109.02707} {Text-to-{Table}: {A} {New}
  {Way} of {Information} {Extraction}}.
\newblock \emph{arXiv preprint}.
\newblock ArXiv:2109.02707 [cs].

\bibitem[{Xing et~al.(2024)Xing, He, Zhou, Dong, Han, Zhang, and
  Chaudhuri}]{xing2024table}
Junjie Xing, Yeye He, Mengyu Zhou, Haoyu Dong, Shi Han, Dongmei Zhang, and
  Surajit Chaudhuri. 2024.
\newblock Table-llm-specialist: Language model specialists for tables using
  iterative generator-validator fine-tuning.
\newblock \emph{arXiv preprint arXiv:2410.12164}.

\bibitem[{Xu and Veeramachaneni(2018)}]{xu_synthesizing_2018}
Lei Xu and Kalyan Veeramachaneni. 2018.
\newblock \href {http://arxiv.org/abs/1811.11264} {Synthesizing {Tabular}
  {Data} using {Generative} {Adversarial} {Networks}}.
\newblock \emph{arXiv preprint}.
\newblock ArXiv:1811.11264 [cs, stat].

\bibitem[{Yang et~al.(2022)Yang, Gupta, Upadhyay, He, Goel, and
  Paul}]{yang-etal-2022-tableformer}
Jingfeng Yang, Aditya Gupta, Shyam Upadhyay, Luheng He, Rahul Goel, and Shachi
  Paul. 2022.
\newblock \href {https://doi.org/10.18653/v1/2022.acl-long.40}
  {{T}able{F}ormer: Robust transformer modeling for table-text encoding}.
\newblock In \emph{Proceedings of the 60th Annual Meeting of the Association
  for Computational Linguistics (Volume 1: Long Papers)}, pages 528--537,
  Dublin, Ireland. Association for Computational Linguistics.

\bibitem[{Yin et~al.(2020)Yin, Neubig, Yih, and Riedel}]{yin2020tabert}
Pengcheng Yin, Graham Neubig, Wen-tau Yih, and Sebastian Riedel. 2020.
\newblock Tabert: Pretraining for joint understanding of textual and tabular
  data.
\newblock \emph{arXiv preprint arXiv:2005.08314}.

\bibitem[{Zha et~al.(2023)Zha, Zhou, Li, Wang, Huang, Yang, Yuan, Su, Li, Su
  et~al.}]{zha2023tablegpt}
Liangyu Zha, Junlin Zhou, Liyao Li, Rui Wang, Qingyi Huang, Saisai Yang, Jing
  Yuan, Changbao Su, Xiang Li, Aofeng Su, et~al. 2023.
\newblock Tablegpt: Towards unifying tables, nature language and commands into
  one gpt.
\newblock \emph{arXiv preprint arXiv:2307.08674}.

\bibitem[{Zhang et~al.(2023)Zhang, Yue, Li, and Sun}]{zhang2023tablellama}
Tianshu Zhang, Xiang Yue, Yifei Li, and Huan Sun. 2023.
\newblock Tablellama: Towards open large generalist models for tables.
\newblock \emph{arXiv preprint arXiv:2311.09206}.

\bibitem[{Zhang et~al.(2020)Zhang, Kishore*, Wu*, Weinberger, and
  Artzi}]{zhang_bertscore_2020}
Tianyi Zhang, Varsha Kishore*, Felix Wu*, Kilian~Q. Weinberger, and Yoav Artzi.
  2020.
\newblock \href {https://openreview.net/forum?id=SkeHuCVFDr} {{BERTScore}:
  {Evaluating} {Text} {Generation} with {BERT}}.
\newblock In \emph{International {Conference} on {Learning} {Representations}}.

\bibitem[{Zhao et~al.(2022)Zhao, Li, Li, and
  Zhang}]{zhao-etal-2022-multihiertt}
Yilun Zhao, Yunxiang Li, Chenying Li, and Rui Zhang. 2022.
\newblock \href {https://doi.org/10.18653/v1/2022.acl-long.454}
  {{M}ulti{H}iertt: Numerical reasoning over multi hierarchical tabular and
  textual data}.
\newblock In \emph{Proceedings of the 60th Annual Meeting of the Association
  for Computational Linguistics (Volume 1: Long Papers)}, pages 6588--6600,
  Dublin, Ireland. Association for Computational Linguistics.

\bibitem[{Zhao et~al.(2023)Zhao, Long, Liu, Nan, Chen, Kamoi, Liu, Tang, Zhang,
  and Cohan}]{zhao_docmath-eval_2023}
Yilun Zhao, Yitao Long, Hongjun Liu, Linyong Nan, Lyuhao Chen, Ryo Kamoi, Yixin
  Liu, Xiangru Tang, Rui Zhang, and Arman Cohan. 2023.
\newblock {DocMath}-{Eval}: {Evaluating} {Numerical} {Reasoning} {Capabilities}
  of {LLMs} in {Understanding} {Long} {Documents} with {Tabular} {Data}.
\newblock \_eprint: 2311.09805.

\bibitem[{Zhao et~al.(2021)Zhao, Kunar, Birke, and Chen}]{zhao2021ctab}
Zilong Zhao, Aditya Kunar, Robert Birke, and Lydia~Y Chen. 2021.
\newblock Ctab-gan: Effective table data synthesizing.
\newblock In \emph{Asian Conference on Machine Learning}, pages 97--112. PMLR.

\bibitem[{Zhu et~al.(2021)Zhu, Lei, Huang, Wang, Zhang, Lv, Feng, and
  Chua}]{zhu-etal-2021-tat}
Fengbin Zhu, Wenqiang Lei, Youcheng Huang, Chao Wang, Shuo Zhang, Jiancheng Lv,
  Fuli Feng, and Tat-Seng Chua. 2021.
\newblock \href {https://doi.org/10.18653/v1/2021.acl-long.254} {{TAT}-{QA}: A
  question answering benchmark on a hybrid of tabular and textual content in
  finance}.
\newblock In \emph{Proceedings of the 59th Annual Meeting of the Association
  for Computational Linguistics and the 11th International Joint Conference on
  Natural Language Processing (Volume 1: Long Papers)}, pages 3277--3287,
  Online. Association for Computational Linguistics.

\end{thebibliography}

\appendix

\section{Appendix} \label{sec:appendix}
\subsection{AI Assistance Acknowledgment}
We acknowledge the use of GitHub Copilot to assist in code completion. 

\subsection{Compute}
All fine-tuning and inference was run on Nvidia A40 GPUs with 48GB GDDR6 memory. Fine-tuning took ~1-2 hours on 8 GPUs in parallel with pytorch distributed data parallel (DDP).

\subsection{Dataset Statistics}
\label{sec:Datasetstats}

Statistics of the text-to-table datasets:

\begin{table}[h]
    \centering
    \begin{tabular}{c c c c}
    \toprule 
        Dataset & Train & Valid & Test  \\
        \midrule
        E2E & 42.1k & 4.7k & 4.7k \\
        WikiTableText & 10k & 1.3k & 2.0k \\
        WikiBio & 582.7k & 72.8k & 72.7k \\
        \bottomrule
    \end{tabular}
    \caption{Statistics of the E2E, WikiTableText, WikiBio, and RotoWire datasets, number of samples across splits}
    \label{tab:ds_statistics}
\end{table}

\subsection{Dataset Examples -- Text to Table}
\label{sec:dataset_examples_t2t}
This section details example textual paragraphs and associated tables from the different datasets. 

\flushleft{\textbf{E2E}: }
\newline The Eagle is a low rated coffee shop near Burger King and the riverside that is family friendly and is less than £20 for Japanese food.
\begin{table}[h]
    \centering
    \begin{tabular}{c|c}
        \toprule 
        Name  & The Eagle \\
        Food  & Japanese \\
        Price range & Less than £20 \\
        Customer Rating & Low \\
        Area & Riverside \\
        Family friendly & Yes \\
        Near & Burger King \\
        \bottomrule
    \end{tabular}
    \label{tab:e2e}
    \vspace{-10pt}
\end{table}

\flushleft{\textbf{WikiTableText}: }
\newline Michelle Schimel was New York State assemblywoman in Portuguese Heritage Society.
\begin{table}[h]
    \centering
    \begin{tabular}{l|l}
    \toprule 
         Title & Potuguese Heritage Society  \\
         Subtitle & Other activities \\
         Name & Michelle Schimel \\
         \bottomrule
    \end{tabular}
    \label{tab:wtt}
    \vspace{-10pt}
\end{table}

\flushleft{\textbf{WikiBio}: } 
\newline Leonard Shenoff Randle (born February 12, 1949) is a former Major League Baseball player. He was the first-round pick of the Washington Senators in the secondary phase of the June 1970 Major League Baseball draft, tenth overall.

\begin{table}[h]
    \centering
    \begin{tabular}{c|c}
       \toprule 
       Debut team  & Washington Senators \\
       Name  & Lenny Randle \\
       Birth Date & 12 February 1949  \\
       \bottomrule
    \end{tabular}
    \label{tab:Wikibio}
    \vspace{-10pt}
\end{table}

\subsection{Dataset Examples -- iTBLS}
\label{sec:dataset_examples_itbls}

\begin{table}[ht]
    \centering
    \begin{tabular}{c|p{0.85\linewidth}}
        \toprule 
          & Example \\
          \midrule 
        1 & What is the 2nd cell value for row 4? \\
        2 & Tell me the final value in the column labeled k \\
        3 & What is the value of the cell to the left of the cell in the bottom right of the table. \\
        \bottomrule 
    \end{tabular}
    \caption{Example interactions in iTBLS \textit{Interpret}}
    \vspace{-10pt}
    \label{tab:interp_example}
\end{table}

\begin{table}[h!]
    \centering
    \begin{tabular}{c|p{0.80\linewidth}}
         \toprule 
         & Example  \\
         \midrule
         1 & The rows 1 and 4 in the Column ``Citation" were accidentally switched. Please rectify the positions of these values so they are where they need to be. \\
         2 & Swap the contents of the second and last cell under repetitions. \\
         3 & Two values in the MCBLp column were put in the reverse spots. I need the values for the FM and PCC rows flipped. \\
         \bottomrule
    \end{tabular}
    \caption{Example interactions in iTBLS \textit{Modify}}
    \label{tab:modif_example}
    \vspace{-10pt}
\end{table}

\begin{table}[h!]
    \centering
    \begin{tabular}{c|p{0.8\linewidth}}
         \toprule 
         & Example \\
         \midrule
         1 & The row 3 of the table shows the values for Peak as 4, X coordinate as 0.100, Y coordinate as -0.150, A as 0.5, standard deviation ($\sigma$) as 0.02, and Local lnZ as -7.824. \\
        2 & The column ``Method 2 (with sub-clustering)" contains the `Nlike' values in different rows: 27,658 in the second row, 69,094 in the third row, 579,208 in the fourth row, and 43,093,230 in the fifth row, while the remaining rows from six to nine contain no data (NaN). \\
        3 & The column R contains eight numerical values in increasing order: 3.34, 3.40, 3.66, 5.06, 6.02, 6.61, 4.05, and 4.11. \\
         \bottomrule
    \end{tabular}
    \caption{Example interactions in iTBLS \textit{Generate}}
    \vspace{-10pt}
    \label{tab:generat_example}
\end{table}

\newpage 
\subsection{Hyperparameters} \label{sec:hyperparams}
Hyperparameters used during training are listed here. 

\begin{table}[htpb]
\addtolength{\tabcolsep}{-1pt}
    \centering
    \begin{tabular}{c|c}
    \toprule
    Parameter & Value \\
    \midrule 
    Rank & 2 \\
    $\alpha$ & 2 \\
    Dropout & 0.01 \\
    Target modules & \texttt{all-linear} \\
    \bottomrule 
    \end{tabular}
    \caption{LoRA Hyperparameters}
    \label{tab:header_gen}
    \vspace{-5pt}
\end{table}

\begin{table}[h]
\addtolength{\tabcolsep}{-1pt}
    \centering
    \resizebox{\columnwidth}{!}{
    \begin{tabular}{ll}
        \toprule
        Parameter & Value \\
        \midrule 
        Learning Rate & 2e-4 \\
        Batch size & 4 \\
        Warmup Schedule & Constant  \\
        Warmup Ratio & 0.03 \\
        Epochs & 5\\
        Optimizer & \texttt{paged\_adamw\_32bit}\tablefootnote{\url{https://huggingface.co/docs/bitsandbytes/main/en/reference/optim/adamw}} \\
        \bottomrule
    \end{tabular}
    }
    \caption{Training Hyperparameters}
    \label{tab:answer_gen}
\end{table}

\subsection{Mechanical Turk Interface}
\label{sec:app_mturk}
\begin{figure*}[h]
    \centering
    \includegraphics[width=0.8\textwidth]{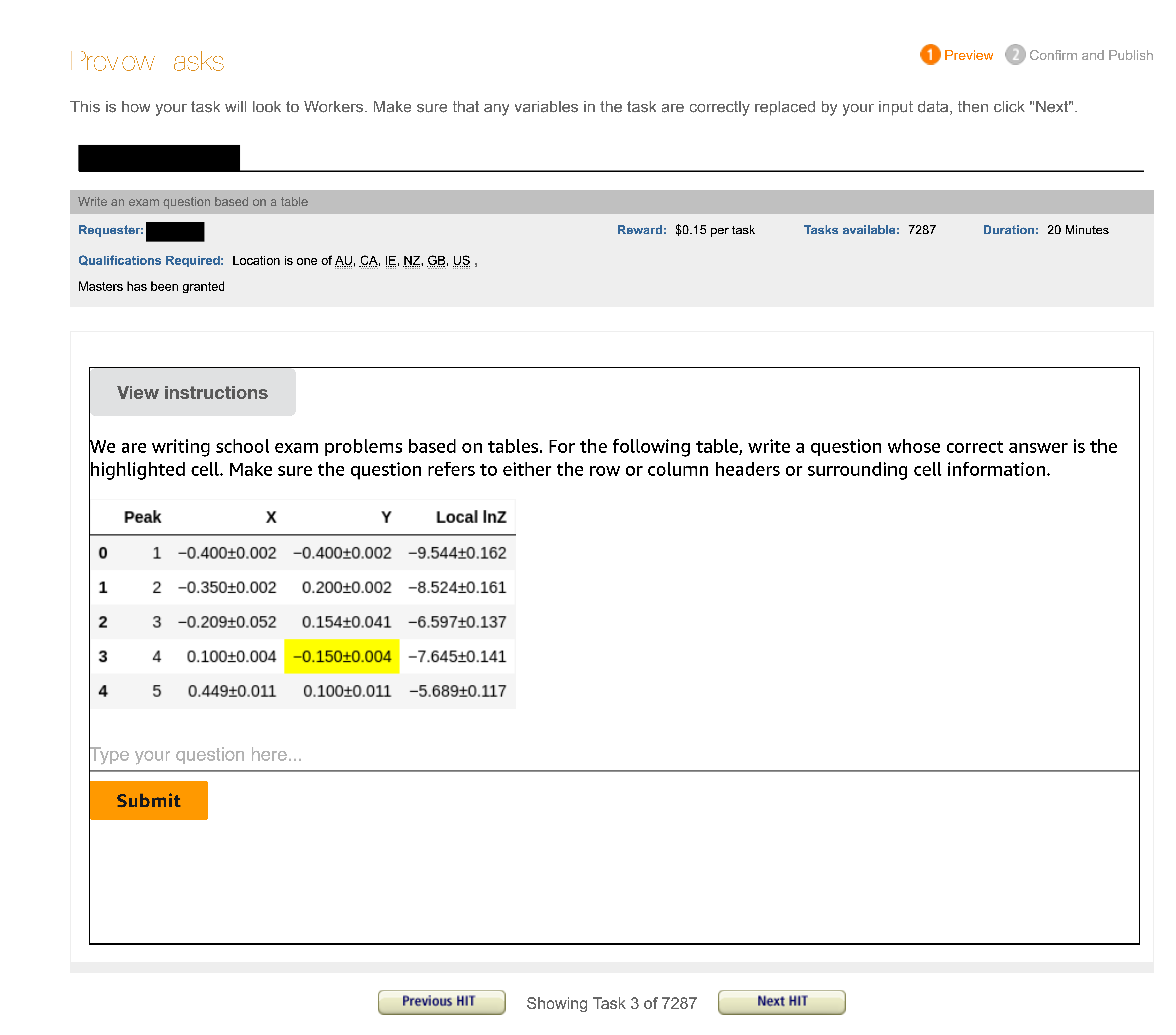}
    \caption{Amazon Mechanical Turk Interface to collect \mbox{iTBLS} interpretation }
    \label{fig:mturk_interp}
\end{figure*}

\begin{figure*}[h]
    \centering
    \includegraphics[width=0.8\textwidth]{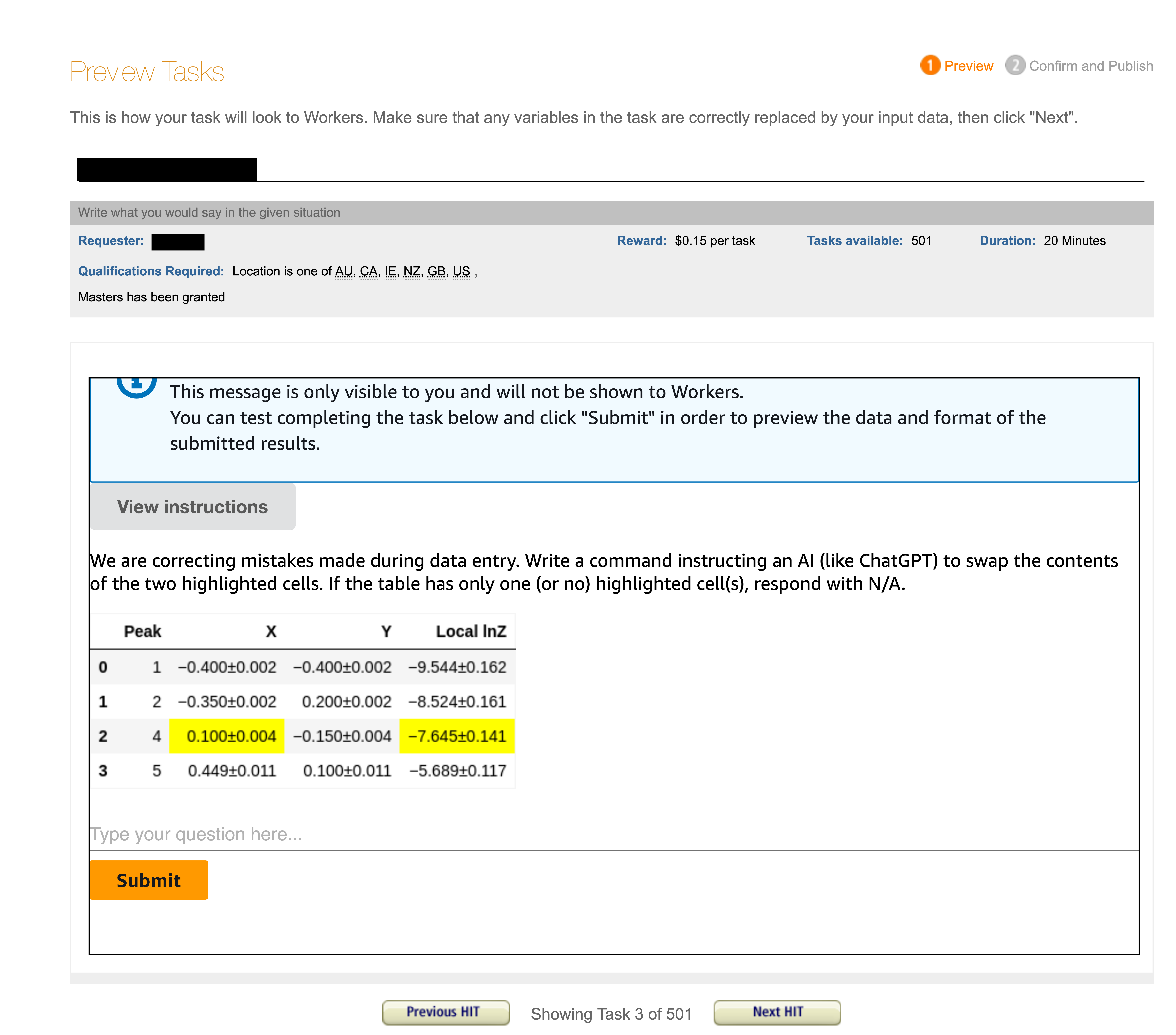}
    \caption{Amazon Mechanical Turk Interface to collect \mbox{iTBLS} modification }
    \label{fig:mturk_modify}
\end{figure*}

\newpage

\section{Example error on the generate task of iTBLS}

\begin{table*}[h]
    \centering
    \textbf{Text}: The column `Standard deviation' contains entries which are both numbers and number sequences: first has 0.45, 0.75, 0, 0.57, second has 0.36, 0.5, 0, 0.34, third is exactly 0, the fourth one is 0.77, fifth is 0.49, and the last one is 0.22. \\
    \vspace{0.5em} 
    {
    \begin{tabular}{c|>{\centering\arraybackslash}p{0.2\textwidth}|>{\centering\arraybackslash}p{0.2\textwidth}|>{\centering\arraybackslash}p{0.2\textwidth}}
    \toprule 
    \toprule 
    \multicolumn{4}{l}{\textbf{Input Table}:} \\ 
    \cmidrule{1-3}
    row ID & Questions & Average score &  \\
    \cmidrule{1-3}
    0 & Q. 1 (a-d) & (3.6 3.93 5 4) & \\
    1 & Q. 2 (a-d) & 4.26 & \\
    2 & Q. 3 & 5 &  \\
    3  & Q. 4 & 3.64 & \\
    4 & Q. 5 & (4.04 4.44 5 4.86) & \\
    5 & GQ & 4.35 & \\
    \midrule 
    \midrule 
    \multicolumn{4}{l}{\textbf{Ground Truth}:} \\ 
    \midrule
    row ID & Questions & Average score & Standard deviation \\
    \midrule
    0 & Q. 1 (a-d) & (3.6 3.93 5 4) & (0.45 0.75 0 0.57) \\
    1 & Q. 2 (a-d) & 4.26 & (0.36 0.5 0 0.34) \\
    2 & Q. 3 & 5 & 0 \\
    3  & Q. 4 & 3.64 & 0.77 \\
    4 & Q. 5 & (4.04 4.44 5 4.86) & \textcolor{blue}{0.49} \\
    5 & GQ & 4.35 & \textcolor{blue}{0.22} \\
    \midrule 
    \midrule 
    \multicolumn{4}{l}{\textbf{Prediction}:} \\ 
    \midrule
    row ID & Questions & Average score & Standard deviation \\
    \midrule
    0 & Q. 1 (a-d) & (3.6 3.93 5 4) & (0.45 0.75 0 0.57) \\
    1 & Q. 2 (a-d) & 4.26 & (0.36 0.5 0 0.34) \\
    2 & Q. 3 & 5 & 0 \\
    3  & Q. 4 & 3.64 & 0.77 \\
    4 & Q. 5 & (4.04 4.44 5 4.86) & \textcolor{red}{(0.49 0.22)} \\
    5 & GQ & 4.35 & \textcolor{red}{0} \\
    \bottomrule
    \end{tabular}
    }
    \caption{Example error for iTBLS generate task. Table source: \url{https://arxiv.org/pdf/1411.4925}}
    \label{tab:difference_answers_itbls_generate}
\end{table*}

\end{document}